\documentclass[runningheads]{llncs}

\usepackage{eccv}

\usepackage{eccvabbrv}

\usepackage{graphicx}
\usepackage{booktabs}
\usepackage{multirow}
\usepackage{multicol}
\usepackage{makecell}
\usepackage{wrapfig}
\usepackage{subcaption}
\usepackage[accsupp]{axessibility}  

\usepackage{hyperref}

\usepackage{orcidlink}

\begin{document}

\title{Few-Shot Open-Vocabulary Remote Sensing Segmentation via Textual Inversion} 

\titlerunning{Few-Shot Open-Vocabulary RS Segmentation via Textual Inversion}

\author{Junhyuk Heo \and
Junghwan Park}

\authorrunning{J.~Heo and J.~Park}

\institute{TelePIX, Seoul, Republic of Korea\\
\email{\{hjh1037, brian897743\}@gmail.com}}

\maketitle

\begin{abstract}
Open-vocabulary segmentation labels arbitrary categories from a text query without per-class training, yet on remote sensing imagery it underperforms on categories it handles reliably elsewhere. We find that much of this gap traces to the text query rather than to the segmentation model. Because these models are not specialized for overhead imagery, the class name that serves as the query is often a weak address into the vision-language embedding space. We show that a better name repairs part of the gap, while the remaining failures call for an address that the tested natural-language rephrasings do not provide. We recover that address from a few examples through textual inversion on a frozen model, keeping inference text only. On a representative benchmark this raises the mean intersection over union on the affected categories from 3.9 to 39.4, and across eight remote sensing datasets it improves over few-shot methods that instead inject visual prompts at inference. The code is available at \href{https://github.com/ROKMC1250/FewShot-OVRS-TI}{https://github.com/ROKMC1250/FewShot-OVRS-TI}
 
\end{abstract}

\begin{figure}[h!]
  \centering
  \includegraphics[width=0.8\textwidth]{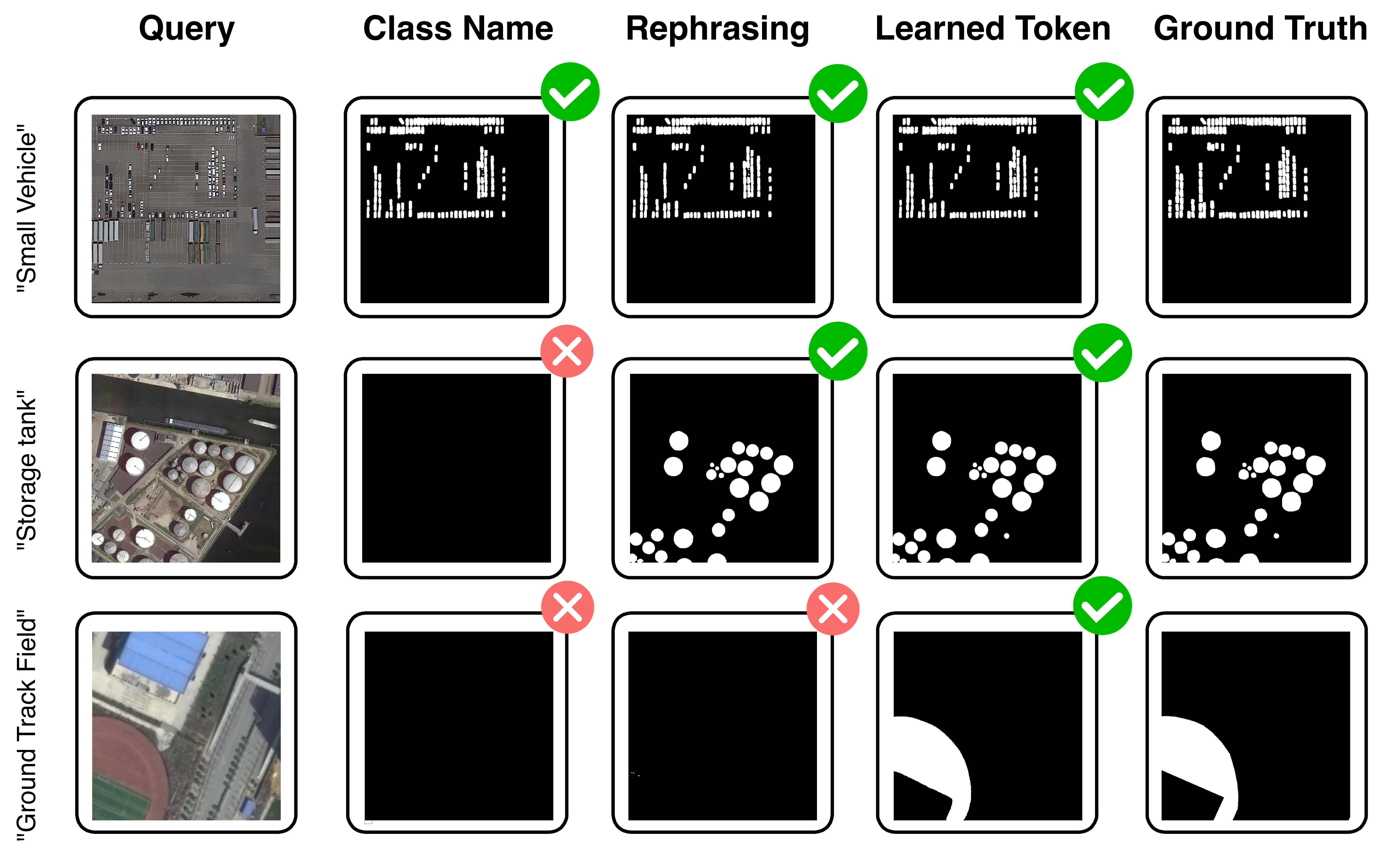}
  \caption{\textbf{Motivation.} Three categories segmented by the same frozen SegEarth-OV3 model under different text queries. Each row is a category and the columns show the query image, the prediction from the class name, the prediction from a better name, the prediction from a token learned on a few support masks, and the ground truth. White marks the predicted region in the prediction columns and the labeled region in the ground truth, and a check marks a query that segments the category while a cross marks one that leaves it nearly empty.}
  \label{fig:motivation}
\end{figure}

\section{Introduction}
 
Open-vocabulary semantic segmentation~\cite{li2022language,liang2023open} labels each pixel by matching a text query against image features in a shared vision-language space, an appealing capability for overhead imagery where the categories of interest are numerous, shifting, and costly to annotate exhaustively. When these models are applied to remote sensing data, they underperform on categories that they handle reliably in everyday photographs. The natural reading is that the segmentation model is the limitation. We find instead that much of the difficulty lies in the text query. A class name acts as an address that points into the shared space, and a segmentation model can only recover a concept that its address locates. Class names are chosen for how a category looks or functions on the ground, so the same word can be a weak address for how the category appears from above, indicating roughly where the category lies without committing the prediction to it and leaving the response faint or nearly empty. This holds even on a strong recent open-vocabulary segmenter built for overhead imagery~\cite{li2025segearth}, the model we build on, so the difficulty is not an artifact of a weak backbone.
 
Consider a category such as a ground track field. Its name describes a function rather than an overhead appearance, the default query segments almost none of it, and choosing the best among plausible alternative names barely helps, since every alternative is still a name and inherits the same difficulty. What helps is to stop naming the category and instead learn its address from a few labeled examples. From a handful of masks we fit a query to which the model commits over the region, and the same image is then segmented well. The examples are used only to fit this query and are discarded afterward, so the category enters inference as a text address in the same form a name would take rather than as visual evidence supplied alongside the query. Figure~\ref{fig:motivation} shows this contrast on three categories, one whose class name already segments it, one that a better name recovers where the class name misses it, and one reached only by a token learned from a few masks.
 
Not every failure has the same cause, and separating the causes is the core of our study. Some categories are merely named awkwardly for the domain, so that a more descriptive name restores them with no visual information, such as a storage tank described by its shape and material. Others, like the example above, are not recovered by renaming and need an address that is learned rather than written. Existing remedies sit unevenly across this divide. Querying with several candidate names and taking the pixelwise maximum over their masks recovers the first kind but leaves the second untouched. A different instinct is to add visual grounding, supplying support masks or points so that the model can locate a category it cannot name~\cite{boudiaf2026segrag,tsai2026few}. This reaches the second kind of failure, but it carries visual evidence into every inference and keeps the category tied to its examples, so the category is segmented without ever becoming a term the text interface can use on its own. It also leaves the cause we identify untouched, correcting nothing about the address the name gave and instead routing around it. What we want to keep is the property that makes open-vocabulary segmentation useful in the first place, a category that can be specified by text and segmented with no examples present at inference. Because the failure we study is a property of the text query, the remedy belongs in the text query, and we show that repairing the address recovers the categories that the tested candidate names do not reach while keeping them nameable.
 
In this work we recover the address itself from a few annotated examples. We keep the segmentation model frozen and learn a small set of pseudo-word embeddings for a category through textual inversion~\cite{galimage}, initialized from the class name and fitted so that the resulting query segments the support. The examples are then set aside, and at inference the category is represented by this learned text address alone, so the model runs as it does for an ordinary name. On remote sensing benchmarks this recovers categories that the tested rephrasings do not reach and improves over few-shot methods that rely on visual prompts, while staying text only throughout, which lets the comparison isolate the effect of a better address.
 
Our contributions are as follows.
 
\begin{itemize}
    \item We reframe open-vocabulary segmentation failure on overhead imagery as a problem of the text query rather than the segmentation model, showing that a better name repairs part of the gap while the remaining failures call for an address that the tested natural-language rephrasings do not provide.
    \item We recover that address from a few annotated examples through textual inversion on a frozen segmentation model, so that at inference the category is segmented from text alone with no examples present.
    \item We show on eight remote sensing datasets that this recovers categories not reached by the tested rephrasings and improves over few-shot methods that inject visual prompts, while keeping inference text only.
\end{itemize}

\begin{figure}[t]
  \centering
  \includegraphics[width=\textwidth]{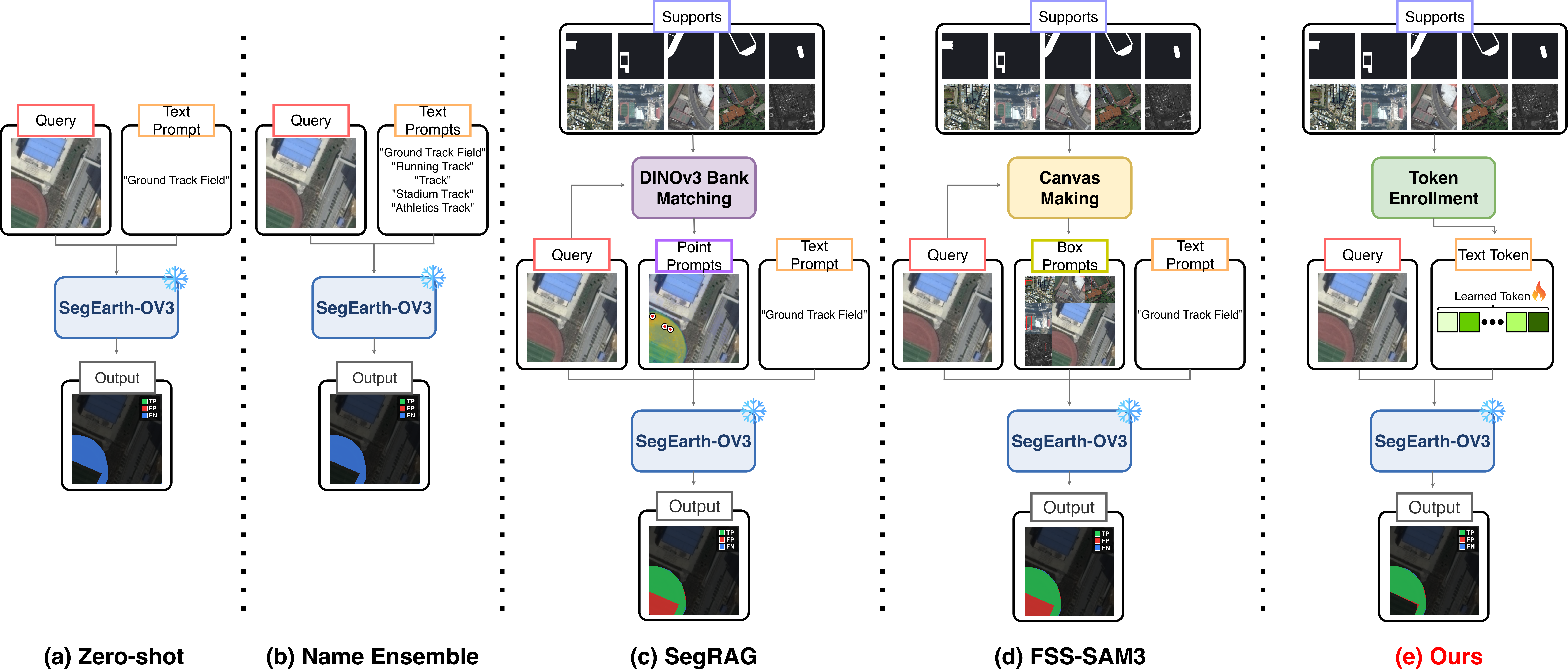}
    \caption{\textbf{Compared approaches.} Five ways to specify a category to the same frozen SegEarth-OV3 model. (a) Zero-shot queries with the class name. (b) Name Ensemble queries with several candidate names, namely synonyms and descriptive phrases, and takes the pixelwise maximum over their masks. (c) SegRAG turns the support masks into point prompts through a DINOv3 feature bank. (d) FSS-SAM3 arranges the support and the query into a single canvas and reads off box prompts. (e) Ours enrolls a learned text token from the support and queries with the token alone, discarding the support. All methods share the frozen model and differ only in how the category enters it. Outputs are colored as true positive, false positive, and false negative.}
  \label{fig:overview}
\end{figure}

\section{Related Work}
 
Our work sits among three lines of research. We build on open-vocabulary semantic segmentation, we share the budget of few-shot segmentation but spend it on the text query rather than on visual prompts, and we belong to a family that adapts the text side of a vision-language model. We review each in turn and mark where our approach departs.
 
\paragraph{\textbf{Open-vocabulary semantic segmentation.}}
Open-vocabulary semantic segmentation labels a pixel by comparing image features with the embedding of a text query in a shared vision-language space, so that categories outside any fixed training set can be named at test time. Early systems align a contrastive image-text encoder~\cite{radford2021learning,jia2021scaling} with dense prediction, either by adapting the encoder for segmentation~\cite{li2022language} or by classifying mask proposals against text~\cite{liang2023open,cho2024cat}. A more recent line builds open-vocabulary labeling on top of a promptable segmentation foundation model, which supplies strong region proposals while a text interface assigns the labels. For overhead imagery this paradigm has been carried into remote sensing through aligned encoders~\cite{liu2024remoteclip} and dedicated open-vocabulary segmenters~\cite{li2025segearth}, the family our backbone belongs to. Across these systems the reported accuracy on remote sensing categories trails what the same models reach on everyday scenes, a gap usually attributed to the visual domain. We instead locate much of it in the text query, and we keep the segmentation model fixed throughout.
 
\paragraph{\textbf{Few-shot segmentation.}}
When a category cannot be specified well in advance, a common remedy is to provide a few labeled examples. Classical few-shot segmentation matches a query image to support masks through learned correlation~\cite{wang2019panet,min2021hypercorrelation}, while in-context approaches segment a query by conditioning on a support pair presented as visual context~\cite{wang2023seggpt,wang2023images}. With promptable foundation models the same idea is realized without training, by turning support regions into spatial prompts such as points or boxes for the query~\cite{zhang2024personalize,liu2023matcher}. The same problem has been studied directly on overhead imagery, where matching and metric learning are adapted to the wide scale variation and dense small objects of aerial scenes~\cite{yao2021scale,wang2021dmml} and where the scarce support is enlarged with imagery synthesized by a diffusion model~\cite{immanuel2025tackling}. Closest to our setting, recent work brings this strategy to the same class of foundation models we use, delivering support evidence as point prompts drawn from a visual feature bank~\cite{boudiaf2026segrag} or as a support and query canvas~\cite{tsai2026few}. What unites these methods is that they reach a novel category by injecting visual evidence at inference. Our approach shares the few-shot budget but spends it differently, recovering a text query that is then used on its own, so that inference stays free of any visual prompt. Figure~\ref{fig:overview} compares these paradigms, from zero-shot naming and renaming to visual few-shot prompting, against our text-only approach.
 
\paragraph{\textbf{Adapting the text side of vision-language models.}}
A second family adapts the language side of a vision-language model rather than supplying images at inference. Prompt learning replaces the hand-crafted template around a class name with a learned context while leaving the name itself fixed~\cite{zhou2022learning,zhou2022conditional}, which tunes how a category is phrased but cannot repair a name that the model resolves only weakly. We instead leave the template alone and learn the name itself, replacing the embedding the name lands on rather than the context around it, so that a category whose name is a weak address is moved rather than rephrased. A related line treats the gap between a source and a target as a transformation in embedding space, learning prompts or projections that move text toward a domain~\cite{yilmaz2024opendas,singha2023ad} or aligning text with visual prototypes to close the modality gap~\cite{liang2022mind,timmermann2025semobridge}. Closest to our method is textual inversion, which represents a concept by a learned pseudo-word embedding optimized on a few examples while the model stays frozen. Introduced for image generation, it has recently been used to address frozen recognition models, including open-vocabulary detection from a handful of examples~\cite{galimage,ruis2025textual}. We differ from this work in three ways. We study open-vocabulary segmentation of remote sensing imagery rather than detection, we tie the learned address to a diagnosis that separates categories a better name repairs from categories not reached by the tested rephrasings, and we keep inference text only by discarding the support once the address is fitted.

\begin{figure}[t]
  \centering

  \begin{subfigure}[t]{0.69\textwidth}
    \centering
    \includegraphics[width=\textwidth]{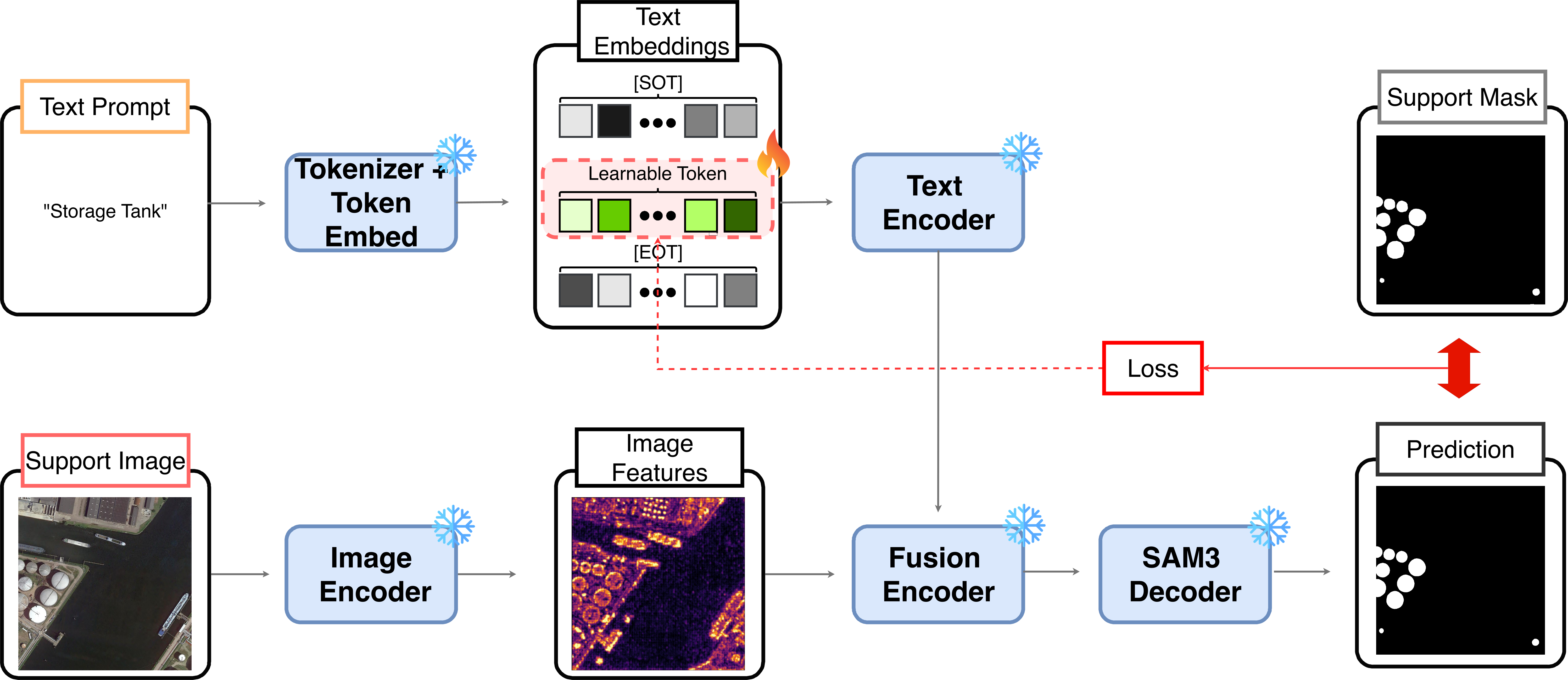}
    \caption{Overview}
    \label{fig:method_overview}
  \end{subfigure}
  \hfill
  \begin{subfigure}[t]{0.29\textwidth}
    \centering
    \includegraphics[width=\textwidth]{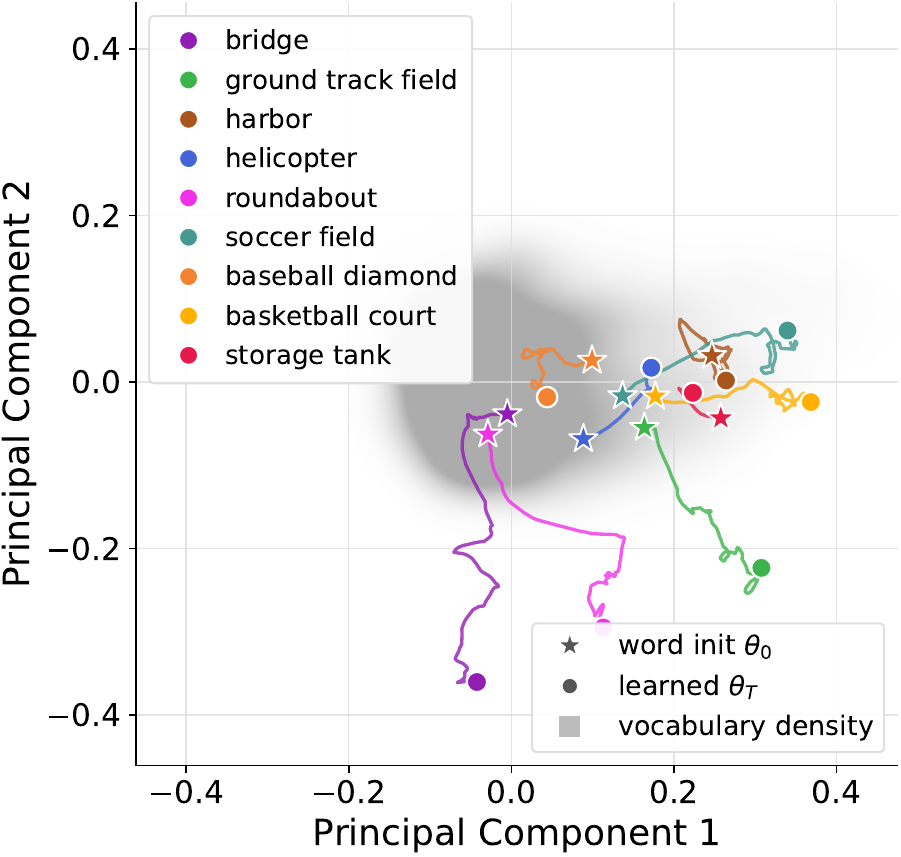}
    \caption{Token trajectory}
    \label{fig:analysis_traj}
  \end{subfigure}

  \caption{
  \textbf{Framework and analysis.}
  The pseudo-word token is optimized while all other modules remain frozen, so gradients flow only to the learnable token through the red dashed path, and the learned token is used as the query at inference without the support mask.
  We further analyze the learned tokens on iSAID under the five-shot setting by projecting the trajectories of the learnable input token embeddings into two dimensions with PCA, where stars indicate the input token embeddings used for name initialization and dots indicate the optimized input token embeddings.
  }
  \label{fig:method_and_analysis}
\end{figure}

\section{Method}
 
We address a category that the tested names do not place well on a frozen open-vocabulary segmentation model. Rather than retrain the model or supply images at test time, we learn a text query for the category from a few support masks and change nothing else, so that the learned query sits in exactly the place a class name would. We keep the model frozen throughout and recover this query from the support, and Figure~\ref{fig:method_overview} shows the pipeline.
 
We write $f$ for the frozen model. For an image $x$ and a category query $q$, the model produces two normalized pixelwise read-outs in $[0,1]$, an instance read-out $I(x,q)$ that responds to individual objects and a semantic read-out $S(x,q)$ that responds to regions, together with a normalized scalar presence score $p(x,q)$ in $[0,1]$ that estimates whether the category appears in the image at all. The score the model assigns to the category at a pixel $u$ gates the stronger of the two read-outs by the presence score,
\begin{equation}
s(x,q)_u = p(x,q)\,\cdot\,\max\!\big(I(x,q)_u,\; S(x,q)_u\big),
\label{eq:score}
\end{equation}
and each pixel is labeled with the category of highest score across the vocabulary, or with background when no category passes a fixed threshold. A category is therefore specified to the model entirely through its query $q$, and the default open-vocabulary choice is to set $q$ to the class name. Given a small support set $\{(x_i, y_i)\}$ of images and binary masks for a category, our aim is to replace that default with a query fitted to the category.
 
\subsection{The Text Query as an Address}
 
A query reaches the model through a fixed sequence of stages. The class name is split into tokens $t_1, \dots, t_N$, each token is read from an embedding table $E$ into an input embedding, the frozen text encoder $g$ maps the resulting sequence to a query representation $z$, a fusion stage $\phi$ conditions the features of a frozen image encoder $h$ on that representation, and a decoder $\psi$ turns the conditioned features into the read-outs above,
\begin{equation}
z = g\big(E(t_1), \dots, E(t_N)\big),
\qquad
\big(I(x,q), S(x,q), p(x,q)\big) = \psi\big(\phi(h(x), z)\big),
\label{eq:pipeline}
\end{equation}
so that the query enters the model only through $z$. Only the early stages depend on the category. The image encoder is the same whatever we are looking for, and the fusion stage and the decoder act on the conditioned features, so the query representation $z$ is the address that selects a category in the shared space, fixed before the model ever sees a particular image. We write address for $z$ throughout and token for an input embedding that $g$ reads, so that a name and a learned pseudo-word differ in the tokens they supply and in nothing that follows.
 
Reaching this address through a name is indirect. The name controls the query only through the embedding table and the frozen encoder, both of which are fixed, so the address a category receives is wherever its name happens to land. For many overhead categories that landing point is a weak one. A name that is rare in the data the encoder was trained on, or that evokes an appearance from the ground rather than from above, produces a query that ranks the region without committing to it, so the response is diffuse, and because every stage after the query is fixed, nothing downstream can sharpen it. A category can thus be segmented well or poorly for reasons that have little to do with the segmentation model and much to do with the address its name provides. We treat this landing point as something to be set from data rather than inherited from the name.
 
\subsection{Learning the Address from Support}
 
We learn the address by textual inversion on the frozen model. We keep the embedding table, the encoders, the fusion stage, and the decoder exactly as they are, and introduce a small set of $M$ learnable input embeddings $v = \{v_1, \dots, v_M\}$ in the same space as the table. These embeddings stand in for the name in the input sequence, so that the frozen text encoder reads them as it would read a word and turns them into the query representation $z = g(v_1, \dots, v_M)$, which places the address under the control of $v$ while every stage that produces it and every stage that follows it stays frozen. Learning the address as a short pseudo-word in the input space, rather than as a free vector inserted after the encoder, keeps the encoder's own transformation in the path and keeps the learned tokens among the inputs the encoder is equipped to interpret. We keep $M$ small because the goal is to relocate the address rather than to describe the category at length, which also makes the address easier to fit from a few masks without absorbing incidental detail of the support images.
 
Given the support set, we fit the embeddings so that the resulting query segments the support, minimizing
\begin{equation}
v^\star = \arg\min_{v}\ \mathcal{L}(v),
\end{equation}
in which every weight of $f$ stays frozen and the gradient reaches only $v$. We build $\mathcal{L}$ from the same normalized read-outs that rank categories at inference, so that the address is optimized for the quantity it will be judged on rather than for a separate feature-space proxy. For a support image we suppress the arguments and write $I_u$, $S_u$ for the two read-outs at a pixel $u$ and $p$ for the presence score. The inference score in Equation~\eqref{eq:score} takes a hard maximum over the read-outs, which would route the gradient to only one of them. During fitting we replace it with the normalized smooth maximum
\begin{equation}
m_\tau(I_u,S_u)
= \tau \log\!\left(\frac{\exp(I_u/\tau)+\exp(S_u/\tau)}{2}\right),
\qquad
\tilde{s}_u = p\,m_\tau(I_u,S_u),
\label{eq:log}
\end{equation}
where $\tau$ is a temperature. The smooth maximum converges to the hard maximum as $\tau$ approaches zero and remains on the same normalized score scale. We use it only in the optimization objective and retain the hard maximum of Equation~\eqref{eq:score} at inference.

Writing $\tilde{s}_i$ for the resulting soft score map over the pixels of support image $x_i$ and $p_i$ for its presence score, the objective is
\begin{equation}
\mathcal{L}(v) = \sum_i \left[\mathcal{L}_{\mathrm{seg}}(\tilde{s}_i,y_i) + \lambda_p\mathcal{L}_{\mathrm{pre}}(p_i)\right] + \lambda_n\mathcal{L}_{\mathrm{neg}}(v),
\label{eq:objective}
\end{equation}
where $\mathcal{L}_{\mathrm{seg}}$ combines binary cross-entropy and soft Dice loss, $\mathcal{L}_{\mathrm{pre}}$ supervises the presence score on positive support images, and $\mathcal{L}_{\mathrm{neg}}$ suppresses the query on support images of the other enrolled categories. The presence term prevents an initially small presence score from attenuating the segmentation gradient, while the cross-category term limits confusion with visually similar categories. The weights $\lambda_p$ and $\lambda_n$ control these two terms.
 
We initialize $v$ from the input embeddings of the class name, which starts the search at the address the name already gives, so that on a category whose name is adequate the optimization has little to do and on a category whose name is weak it is free to move as far as the support requires. Two further choices stabilize the fit from a few masks. We take a slow moving average of $v$ across optimization steps as the final query, which limits the pull of any single support image and yields an address tuned to the category rather than to the particular examples seen, and we augment the support with the eight rotations and reflections that map a square to itself, so that the address does not depend on the orientation of the few masks available, caching the encoder features so that this augmentation adds little cost. A minimum area on the support masks keeps degenerate regions out of the fit. We replace the name in the query with the learned embeddings, and we treat appending them to the name instead as a design choice examined in our experiments.
 
\subsection{Text-Only Inference}
 
The support is used only to fit the address and is then discarded. At test time the category is represented by the learned embeddings $v^\star$ alone, which enter the frozen text encoder in place of a name and give the address $g(v^\star)$, and the model segments a new image exactly as it does for an ordinary text query. Because a learned address and an ordinary name produce a query in the same representation and are scored by the same rule, the enrolled categories and the categories left to their names compete directly in one full-vocabulary assignment, with no separate calibration or rescaling between them. No support image, mask, point, or box is present at inference, so the category travels entirely as a learned text address and the model runs in its original open-vocabulary mode.
 
This property is what lets our results speak to the diagnosis. A method that supplied the support at test time could improve for two different reasons, because it found a better address or because it added visual evidence that the text never carried, and the two would be hard to tell apart. By discarding the support and keeping only the address, we hold the inference interface fixed at a single text query, so that any gain over the name has to come from the address itself. Because the image encoder stays frozen and the learned embeddings enter only the query path, they cannot alter the features the model extracts from an image, so a learned address acts only by selecting and conditioning features that are already present rather than by carrying new visual evidence into inference.
\section{Experiments}

\begin{table*}[t]\centering
\renewcommand{\arraystretch}{1.2}
\caption{\textbf{Quantitative results.} mIoU in percent across eight remote sensing datasets, as
mean and standard deviation over three support draws chosen by seed. SegRAG, FSS-SAM3, and Ours are reported at one, five and ten shots, while Zero-shot, Name Ensemble, and the fully supervised reference do
not depend on shots. Avg is the mean over the eight datasets. The fully supervised reference is a
SegFormer-b0~\cite{xie2021segformer} trained on full data, with values taken from
SegEarth-OV3~\cite{li2025segearth}. Among the few-shot methods the best in each column at each shot count is bold and the second underlined.}
\label{tab:cross_dataset}
\resizebox{\textwidth}{!}{%
\begin{tabular}{lcccccccccc}
\toprule
Method & Shots & iSAID & LoveDA & OpenEarthMap & Potsdam & VDD & UAVid & Vaihingen & UDD5 &
\textbf{Avg} \\
\midrule
Zero-shot & 0 & 27.4 & 47.4 & 44.1 & 57.9 & 64.5 & 58.6 & 60.8 & 71.7 & 54.0 \\
Name Ensemble & 0 & 31.0 & 40.9 & 39.4 & 55.5 & 64.5 & 58.6 & 62.4 & 59.5 & 51.5 \\
\midrule
SegRAG & 1 & 31.5$\pm$1.9 & \underline{38.6$\pm$1.1} & \underline{41.6$\pm$0.3} & 55.9$\pm$0.7 & 63.8$\pm$0.3
& \underline{58.9$\pm$0.7} & \underline{61.9$\pm$1.2} & \underline{63.4$\pm$0.3} & \underline{52.0} \\
FSS-SAM3 & 1 & \underline{31.8$\pm$0.7} & \textbf{42.8$\pm$1.0} & 34.7$\pm$1.2 & \textbf{60.0$\pm$0.4} &
\underline{64.7$\pm$2.2} & 55.9$\pm$4.8 & \textbf{63.4$\pm$1.3} & 56.5$\pm$3.0 & 51.2 \\
\textbf{Ours} & 1 & \textbf{32.1$\pm$1.9} & 33.4$\pm$1.0 & \textbf{44.0$\pm$2.4} &
\underline{56.7$\pm$4.6} & \textbf{65.2$\pm$8.3} & \textbf{65.3$\pm$1.1} & 61.9$\pm$1.0 &
\textbf{66.0$\pm$2.5} & \textbf{53.1} \\
\midrule
SegRAG & 5 & \underline{38.2$\pm$1.5} & 39.9$\pm$0.4 & \underline{44.1$\pm$1.1} & 56.6$\pm$1.2 &
63.9$\pm$0.7 & 59.3$\pm$0.4 & \underline{64.4$\pm$1.0} & \underline{65.2$\pm$1.4} & 54.0 \\
FSS-SAM3 & 5 & 37.3$\pm$0.2 & \textbf{45.7$\pm$2.0} & 37.7$\pm$0.3 & \textbf{65.3$\pm$0.6} &
\underline{65.4$\pm$1.8} & \underline{60.1$\pm$1.0} & \textbf{67.0$\pm$1.0} & 59.6$\pm$1.5 &
\underline{54.8} \\
\textbf{Ours} & 5 & \textbf{42.7$\pm$1.7} & \underline{42.9$\pm$3.7} & \textbf{53.0$\pm$2.9} &
\underline{64.6$\pm$1.4} & \textbf{75.8$\pm$2.4} & \textbf{69.6$\pm$0.3} & 63.8$\pm$1.6 &
\textbf{70.6$\pm$0.7} & \textbf{60.4} \\
\midrule
SegRAG & 10 & \underline{39.9$\pm$0.5} & 41.1$\pm$0.5 & \underline{44.3$\pm$2.3} & 57.7$\pm$1.2 &
\underline{65.8$\pm$1.5} & 61.2$\pm$0.7 & 65.9$\pm$1.3 & \underline{62.6$\pm$1.5} &
\underline{54.8} \\
FSS-SAM3 & 10 & 37.7$\pm$1.5 & \underline{45.6$\pm$0.5} & 35.2$\pm$1.1 & \underline{65.1$\pm$1.1} &
65.1$\pm$3.1 & \underline{62.3$\pm$2.4} & \textbf{67.3$\pm$1.9} & 59.5$\pm$0.8 & 54.7 \\
\textbf{Ours} & 10 & \textbf{43.2$\pm$1.2} & \textbf{48.3$\pm$0.8} & \textbf{55.0$\pm$2.0} &
\textbf{65.5$\pm$0.5} & \textbf{76.7$\pm$0.1} & \textbf{68.7$\pm$0.5} & \underline{67.1$\pm$0.8} &
\textbf{71.3$\pm$0.6} & \textbf{62.0} \\
\midrule
Fully supervised & -- & 36.2 & 50.0 & 64.4 & 74.3 & 62.9 & 59.7 & 61.2 & 56.5 & 58.2 \\
\bottomrule
\end{tabular}}
\end{table*}

We evaluate the learned address on iSAID~\cite{waqas2019isaid} and seven further remote sensing datasets, with the comparison across datasets as the main result and iSAID as a representative case that carries the diagnosis. We compare against a zero-shot lower bound, a renaming baseline, two few-shot methods that inject visual prompts at inference, and a fully supervised reference. The comparison holds the frozen segmentation backbone, vocabulary, full-vocabulary assignment rule, and evaluation protocol fixed while varying how category-specific information enters the model. This lets us compare a text-only learned address with methods that carry visual evidence into inference without attributing the difference to the underlying segmenter or evaluation rule. We then examine where the learned query moves and what it changes in the model's matching response.

\begin{table*}[t]\centering
\renewcommand{\arraystretch}{1.}
\caption{\textbf{Per-class results on iSAID.} Per-class IoU in percent at five support masks, computed for each of the five categories whose zero-shot IoU is below ten on a held-out selection split, as mean and standard deviation over three support draws chosen by seed. Mean is the mean over the five categories. Among the methods the best in each column is bold and the second underlined.}
\label{tab:isaid_perclass}
\footnotesize
\setlength{\tabcolsep}{3pt}
\resizebox{\textwidth}{!}{%
\begin{tabular}{lcccccc}
\toprule
Method & \makecell{Storage\\tank} & \makecell{Baseball\\diamond} & \makecell{Basketball\\court} & \makecell{Ground\\track field} & \makecell{Soccer\\field} & Mean \\
\midrule
Zero-shot & 2.8 & 3.1 & 7.5 & 0.4 & 5.5 & 3.9 \\
Name Ensemble & \underline{61.3} & 3.8 & 6.1 & 3.7 & 14.6 & 17.9 \\
SegRAG & 51.9$\pm$6.5 & \textbf{45.8$\pm$2.2} & \underline{13.6$\pm$3.4} & 6.5$\pm$0.9 & \textbf{37.9$\pm$14.4} & \underline{31.1$\pm$1.8} \\
FSS-SAM3 & 31.7$\pm$15.6 & \underline{42.0$\pm$1.9} & 11.5$\pm$2.3 & \underline{7.0$\pm$2.4} & \underline{32.5$\pm$8.4} & 24.9$\pm$0.3 \\
\cmidrule{1-7}
\textbf{Ours} & \textbf{71.6$\pm$5.4} & 36.9$\pm$8.1 & \textbf{33.0$\pm$10.1} & \textbf{33.6$\pm$0.6} & 21.9$\pm$9.7 & \textbf{39.4$\pm$3.2} \\
\bottomrule
\end{tabular}}
\end{table*}

\subsection{Implementation Details}
 
All methods share a frozen SegEarth-OV3~\cite{li2025segearth}, a SAM3-based open-vocabulary segmenter for overhead imagery, and the same full-vocabulary argmax, differing only in how the support enters the query. The zero-shot lower bound queries with the class name, and Name Ensemble adds ten candidate names per category, synonyms and descriptive phrases generated by a language model on top of the names SegEarth-OV3 already uses, and takes the pixelwise maximum over their masks, so it measures how far better naming alone can reach without using any support. The visual few-shot methods extract support features with a frozen DINOv3 encoder~\cite{simeoni2025dinov3} and deliver them as point prompts from a prototype bank for SegRAG~\cite{boudiaf2026segrag} or as a single support and query canvas for FSS-SAM3~\cite{tsai2026few}. Besides iSAID we evaluate on LoveDA~\cite{wang2021loveda}, OpenEarthMap~\cite{xia2023openearthmap}, Potsdam and Vaihingen~\cite{sohn2012isprs}, UDD5~\cite{chen2018large}, VDD~\cite{cai2025vdd}, and UAVid~\cite{lyu2020uavid}. We enroll every foreground category on these seven datasets. For iSAID, we partition the validation data into a held-out selection split and a disjoint support pool. We compute zero-shot per-class IoU only on the held-out selection split, select the five categories below ten, and fix this category set before drawing any support masks. The remaining categories continue to use their class names. We report one, five and ten support masks and average every number over three seeds. Our method learns two pseudo-word tokens per category and queries with the tokens in place of the name. We fit the tokens with Adam at a learning rate of $10^{-2}$ for 300 steps and take an exponential moving average with decay 0.99 as the final query.
We set the temperature $\tau$ in Equation~\eqref{eq:log} to $0.1$ and weight the presence and the cross-category terms with $\lambda_p$ at $0.5$ and $\lambda_n$ at $0.5$, and we hold these three values fixed across every dataset and every shot count rather than tuning them per category.
A pixel falls to background when no category passes a fixed score threshold, and we keep this threshold at the value SegEarth-OV3 sets for each dataset, namely $0.5$ on iSAID and LoveDA, $0.3$ on VDD and UAVid, and $0.1$ on OpenEarthMap, Potsdam, Vaihingen, and UDD5, so that every method in our comparison labels background by the same rule and none of them tunes it.
Support masks are drawn from the training split on the seven additional datasets and from the disjoint support pool on iSAID. For a category we sample uniformly at random among the patches in which it covers at least fifty pixels and at least one percent of the patch area, falling back to the largest such patches when fewer of them exist than the shot count. Each of the three seeds gives an independent draw. The official evaluation split is never used for category selection or support sampling, and every number we report is measured on that split with every patch that supplied a support mask removed, so that no support image enters the evaluation. We run all experiments on four NVIDIA RTX A6000 GPUs. 

\subsection{Main Results}

\begin{figure}[t]
\centering
\includegraphics[width=\linewidth]{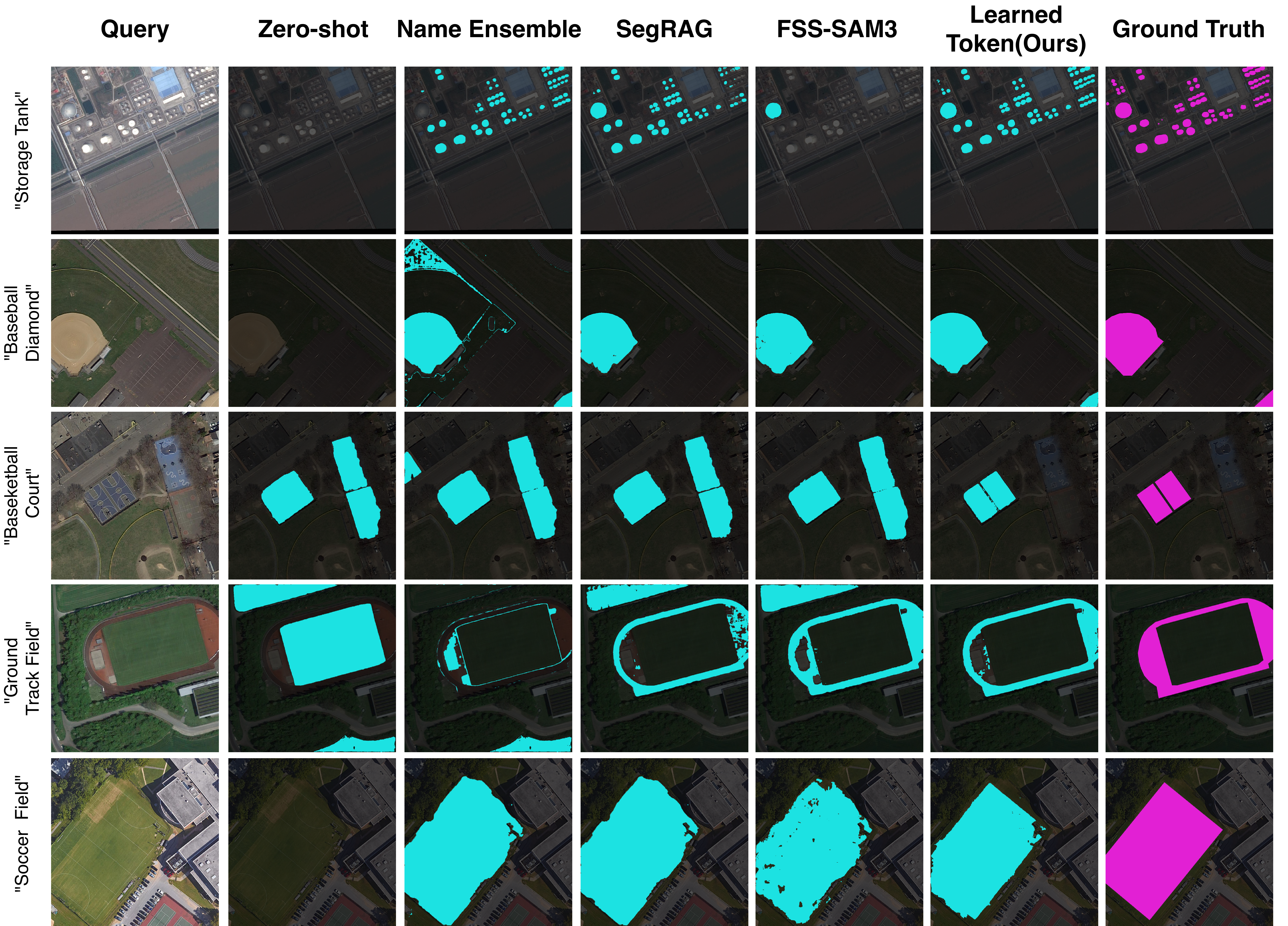}
\caption{\textbf{Qualitative comparison.} Qualitative comparison on iSAID under full-vocabulary argmax. The columns show the input, the zero-shot name query, the Name Ensemble query, a visual few-shot baseline, the learned token, and the ground truth. The rows show the categories whose zero-shot IoU is below ten.}
\label{fig:qualitative}
\end{figure}

Across the eight datasets the learned address gives the best average among the few-shot methods at every shot count, reaching 60.4 at five shots and 62.0 at ten, ahead of the zero-shot lower bound at 54.0 and of both visual few-shot methods, as Table~\ref{tab:cross_dataset} reports. Setting iSAID aside, where only five of the sixteen categories are enrolled, the gains over zero-shot are largest on VDD, OpenEarthMap, UAVid, and Potsdam, and the average improves across datasets of different sensors and label sets, which indicates the recovery is a property of the domain rather than of one benchmark. The address does not exceed the zero-shot baseline on LoveDA and UDD5 at five shots, and LoveDA recovers at ten. Name Ensemble behaves differently, since a loosely fitting name draws a false response that the full-vocabulary argmax then assigns elsewhere, so it falls below the zero-shot baseline on average, while the learned address improves on it at five and ten shots and stays within a point of it at one. The table also lists a fully supervised SegFormer-b0~\cite{xie2021segformer} as a reference, with values taken from SegEarth-OV3~\cite{li2025segearth}.

\setlength{\intextsep}{0pt}
\begin{wrapfigure}[19]{r}{0.45\linewidth}
\centering
\includegraphics[width=\linewidth]{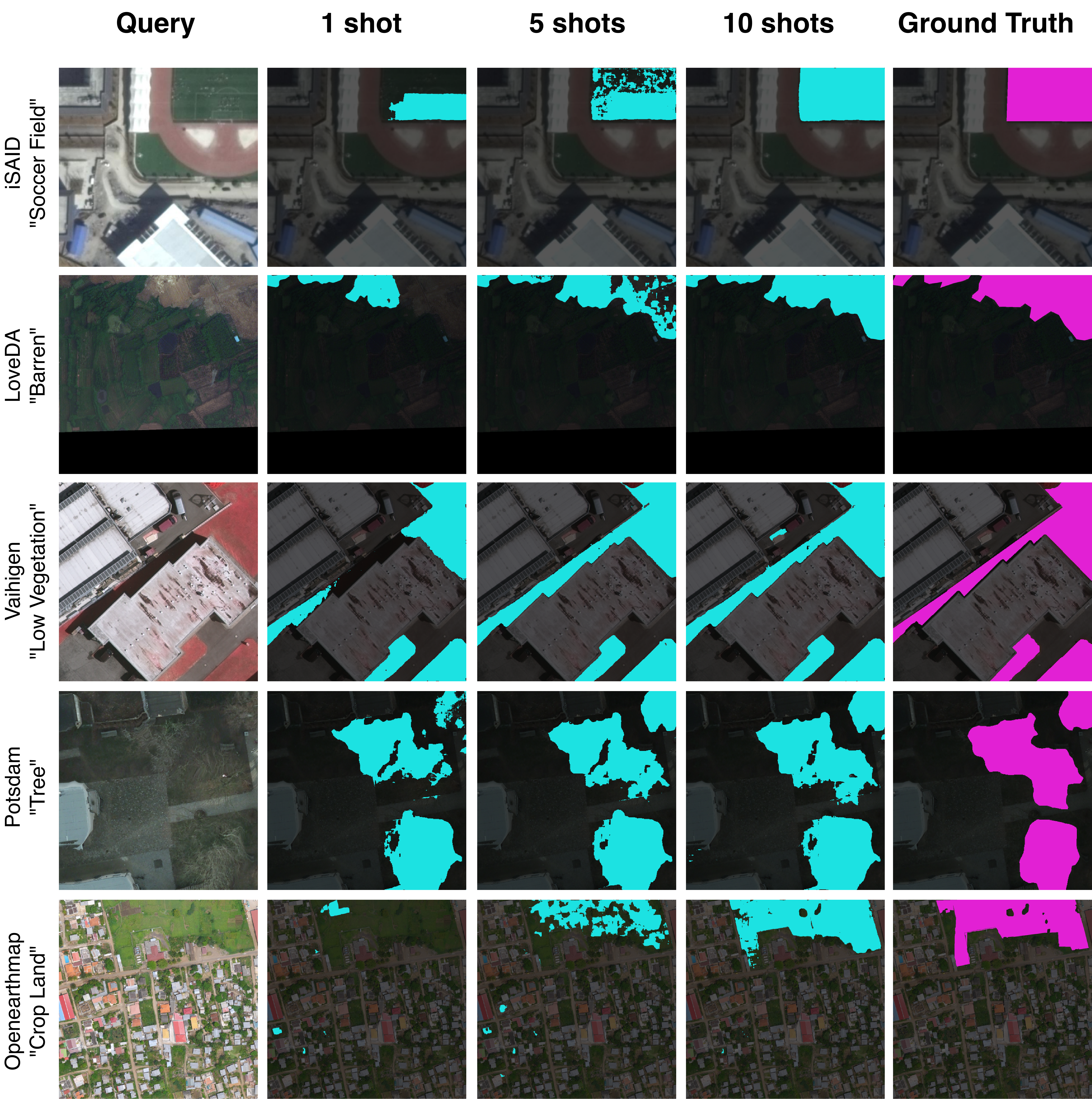}
\caption{\textbf{Effect of support size.} Predictions of the learned token as the number of support masks grows, shown in cyan against the ground truth in magenta.}
\label{fig:shot_results}
\end{wrapfigure} On iSAID, where open-vocabulary segmentation fails on categories it handles well in everyday scenes, enrolling the five categories whose zero-shot IoU is below ten recovers them while leaving the rest in place. The full sixteen-way mIoU rises from 27.4 to 42.7 at five shots and the mean over the enrolled categories rises from 3.9 to 39.4, as Table~\ref{tab:cross_dataset} and Table~\ref{tab:isaid_perclass} report. The learned address has the best enrolled mean and, among the few-shot methods, is strongest on three of the five categories, reaching 71.6 on storage tank, 33.0 on basketball court, and 33.6 on ground track field.

The per-class results separate the two kinds of failure the introduction describes. Storage tank is already recovered by a better name under Name Ensemble, from 2.8 to 61.3, so its failure was a matter of naming. Ground track field is barely moved by renaming, from 0.4 to 3.7, and is reached only by the learned token at 33.6, so no name in the candidate set locates it and the address has to be learned. The visual methods keep an edge on baseball diamond and soccer field, marking the boundary where the difficulty is visual rather than nominal.

Figure~\ref{fig:qualitative} shows these patterns under full-vocabulary argmax. The zero-shot name leaves most of these categories empty or mislocated. Name Ensemble recovers a category that a description already reaches, and on storage tank its mask comes close to the one the learned token produces, while it leaves ground track field much as the name does. The learned token recovers both kinds and tends to a more compact mask than the visual baselines, which over-extend on basketball court.

The learned address also strengthens as the support grows, as Figure~\ref{fig:shot_results} traces across one category from each of five datasets, namely soccer ball field on iSAID, barren on LoveDA, low vegetation on Vaihingen, tree on Potsdam, and crop land on OpenEarthMap. Reading the columns from left to right, the query image is followed by the prediction at one, five, and ten support masks and then by the ground truth. With a single mask the address already places the category in roughly the right region, and as more masks are added the prediction fills the category more completely and sheds the scattered response a single example leaves behind, so that at ten masks it aligns most closely with the ground truth. This follows the rise in average accuracy from one to ten shots in Table~\ref{tab:cross_dataset}, where the learned address climbs from 53.1 to 60.4 and then to 62.0. The gain is not uniform, since a category whose overhead appearance is hard to separate is reached only partly even at ten masks, but a larger support consistently moves the prediction toward the labeled region rather than away from it.

\setlength{\intextsep}{0pt}
\begin{wrapfigure}[18]{r}{0.45\linewidth}
\centering
\includegraphics[width=\linewidth]{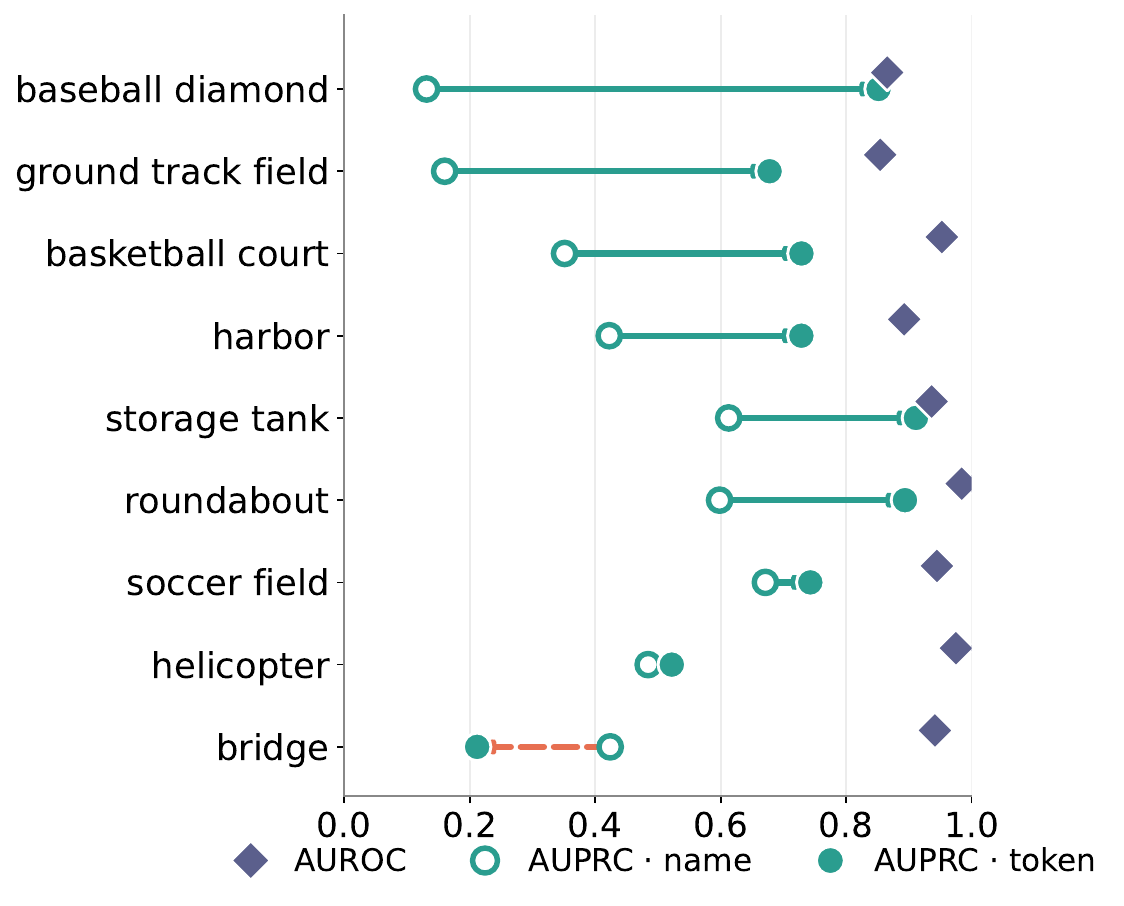}
\caption{\textbf{Ranking and precision.} Analysis on iSAID at five shots. For nine categories, the AUROC of the name query and the AUPRC of the name query and of the learned token, against the ground truth.}
\label{fig:analysis_auprc}
\end{wrapfigure}

\subsection{Where the Address Moves and What It Changes}

For this analysis we enroll a token for nine iSAID categories rather than the five used for the main result, so that the rank correlation rests on enough points. We read the model's own matching response against the ground truth with two threshold-free measures, the area under the ROC curve for where the response ranks the target region and the area under the precision-recall curve for how concentrated that response is. We look at both because a foreground category covers a small fraction of the pixels, so a query can rank the region well and still place little of its response on it.

The model, the image, and the decoder stay fixed while only the query is learned, so any change comes from the query alone. Figure~\ref{fig:analysis_traj} projects the learnable input token embeddings for each category from their name initialization through optimization, and the tokens leave the initialization and settle at a separate location rather than refining it in place.
 
Figure~\ref{fig:analysis_auprc} reads what the relocation changes in the matching response. The name already ranks the ground-truth region well, with a high area under the ROC curve even where the IoU is near zero, so the features separate the region and the failure is not one of ranking. What the name lacks is precision, and the learned token raises the area under the precision-recall curve in step with the IoU, with a Spearman rank correlation of 0.77 between the two changes at a p-value of 0.016. The token raises the area under the precision-recall curve on eight of the nine categories and lowers it only on bridge. Because the features are fixed and only the query changed, the precision the name lacked was recoverable from the query alone, and because precision is read here without a threshold, a per-category threshold on the name cannot reproduce the gain.

\subsection{Ablation Study}

\begin{table*}[t]\centering
\renewcommand{\arraystretch}{1.}
\caption{\textbf{Ablation on iSAID.} Per-class IoU at five support masks over the five enrolled categories, reported as the mean over the same three support draws used in Table~\ref{tab:isaid_perclass}. The final column is the mean over the five categories. The upper block varies the prompt form, where we replace the name with the learned tokens, append the tokens after the name, or replace the name with tokens initialized at random. The lower block varies the number of tokens under the adopted replace form.}
\label{tab:ablation}
\setlength{\tabcolsep}{4pt}
\begin{tabular}{lcccccc}
\toprule
Setting & \makecell{Storage\\tank} & \makecell{Baseball\\diamond} & \makecell{Basketball\\court} & \makecell{Ground\\track field} & \makecell{Soccer\\field} & Mean \\
\midrule
Replace name   & 71.6 & 36.9 & 33.0 & 33.6 & 21.9 & 39.4 \\
Append to name & 72.6 & 19.8 & 38.5 & 26.4 & 23.2 & 36.1 \\
Random init    & 72.0 & 15.6 & 17.7 & 37.4 & 3.9 & 29.3 \\
\midrule
1 token  & 69.5 & 34.0 & 39.4 & 34.9 & 28.0 & 41.2 \\
2 tokens & 71.6 & 36.9 & 33.0 & 33.6 & 21.9 & 39.4 \\
4 tokens & 70.8 & 16.5 & 44.6 & 41.5 & 37.7 & 42.2 \\
8 tokens & 70.2 & 14.9 & 42.6 & 38.9 & 42.5 & 41.8 \\
\bottomrule
\end{tabular}
\end{table*}

We adopt the form that replaces the name with the learned tokens, initialized from the name, and Table~\ref{tab:ablation} reports each setting per category over the same three support draws used in the main comparison. Among the prompt-form variants, replacing the name gives the best mean at 39.4, ahead of appending the tokens to the name at 36.1 and random initialization at 29.3. Random initialization is the weakest on average and collapses on individual categories such as baseball diamond and soccer field, while initialization from the name gives a more stable starting point. We replace the name rather than append to it because replacement makes the category a self-contained learned term that no longer depends on the original name, which is the property we want to return to the text interface. The lower block varies the number of tokens under this form. The average is not monotonic in the token count, and the best count varies across categories. Four tokens gives the highest mean in this ablation, but its gain is concentrated in basketball court, ground track field, and soccer field while its baseball diamond result falls sharply. We therefore use two tokens as a small fixed budget that keeps the learned term compact and avoids selecting the token count by category. An exponential moving average of the tokens gives a slightly steadier result than the final step. Per-class accuracy varies across support draws, for example a standard deviation near ten points on soccer field, which is why we average every result over three seeds.

\subsection{Enrollment and Inference Cost}

Enrolling a category is a one-time cost, since the fit runs the 300 optimization steps on the support and the cached encoder features keep the dihedral augmentation from multiplying this cost. Once enrolled, the learned tokens enter the text encoder exactly as a name would, the address is computed once and reused across every image, and no support image, mask, or feature bank stays in memory, so the model runs at the speed and memory footprint of the zero-shot baseline. The visual few-shot methods instead carry a prototype bank or a support canvas into every inference, so their cost recurs with each image while ours is paid once at enrollment.

\section{Limitations and Future Work}

Our diagnosis separates categories whose failure is nominal from categories whose failure is visual, and the learned address speaks only to the first. Where the difficulty lies in how a category appears from above, as on baseball diamond and soccer field, an address fitted from a few masks recovers less than a method that carries visual evidence into inference, since no relocation of the query can supply detail the frozen features do not separate. The address is also fitted per category and held fixed across a dataset, which leaves it sensitive to the particular masks seen and blind to appearance shifts a single embedding cannot follow.

These limits point to the text query and the visual prompt as complementary rather than competing. A category could be routed to a learned address when its failure is nominal and to visual grounding when it is visual, and the fixed address could give way to one conditioned on the input image while still entering inference as text. For a practitioner, enrollment amortizes a single fit over every later text query on an archive, while visual prompting stays the better tool when the support must remain in the loop.

\section{Conclusion}
 
In this work we traced the failure of open-vocabulary segmentation on overhead imagery to the text query rather than to the segmentation model. A class name is a weak address into the shared vision-language space, one that ranks the target region without committing to it. We recovered the address from a few support masks through textual inversion on a frozen model, and we kept inference text only by discarding the support once the address was fitted. Across eight remote sensing datasets the learned address improves on the zero-shot name and on few-shot methods that instead inject visual prompts at inference, and the per-class results separate the failures a better name repairs from those that renaming does not reach. Where existing few-shot remedies carry visual evidence into every inference, our approach corrects the query itself and returns the category to the text interface as a learned term. This suggests that much of the remote sensing gap is a problem of naming rather than of seeing, and that an address fitted from a few masks and applied with no support present is a simple way to narrow it.

\clearpage
%
%
\bibliographystyle{splncs04}
\bibliography{main}
\end{document}